\documentclass[letterpaper, 9pt, conference]{ieeeconf}  

\IEEEoverridecommandlockouts                              

\newcommand{\eg}{\textit{e}.\textit{g}.}

\newcommand{\ie}{\textit{i}.\textit{e}.}

\usepackage{graphicx}
\usepackage{amsmath}
\usepackage{amssymb}
\usepackage{booktabs}
\usepackage{multirow}
\usepackage{url}
\usepackage{xcolor}
\usepackage{color}
\usepackage{pifont}
\usepackage{hyperref}
\hypersetup{pdfstartview=FitH,
            colorlinks=true,
            linkcolor=red,
            anchorcolor=blue,
            citecolor=black
            }

\title{\LARGE \bf
Chasing Day and Night: Towards Robust and Efficient All-Day Object Detection Guided by an Event Camera
}

\author{Jiahang Cao$^{1}$, Xu Zheng$^{1\diamondsuit}$, Yuanhuiyi Lyu$^{1\diamondsuit}$, Jiaxu Wang$^{1}$, Renjing Xu$^{1\dagger}$, Lin Wang$^{1,2\dagger}$
\thanks{This work was sponsored by Zhejiang Lab Open Research Project (NO.K2022PH0AB01).}
\thanks{$^\dagger$Corresponding authors; $^\diamondsuit$Co-second authors}
\thanks{$^{1}$Jiahang Cao and Jiaxu Wang are with MICS Thrust,
        HKUST(GZ),
        \newline Email: {\tt\small jcao248@connect.hkust-gz.edu.cn, \newline jwang457@connect.hkust-gz.edu.cn}}%
\thanks{$^{1}$Xu Zheng and Yuanhuiyi Lyu are with AI Thrust,
        HKUST(GZ),
        \newline Email: {\tt\small xzheng287@connect.hkust-gz.edu.cn,\newline yuanhuiyilv@hkust-gz.edu.cn}}%
\thanks{$^{1}$Renjing Xu is with MICS Thrust,
        HKUST(GZ), Guangzhou, China,
        Email: {\tt\small renjingxu@ust.hk}}%
\thanks{$^{1,2}$Lin Wang is with AI Thrust, HKUST(GZ), Guangzhou, and Dept. of CSE, HKUST, Hong Kong SAR, China,
        Email: {\tt\small linwang@ust.hk}}%
\\
}

\begin{document}

\maketitle
\thispagestyle{empty}
\pagestyle{empty}


\begin{abstract}
The ability to detect objects in all lighting (\ie, normal-, over-, and under-exposed) conditions is crucial for real-world applications, such as self-driving. Traditional RGB-based detectors often fail under such varying lighting conditions. 
Therefore, recent works utilize novel event cameras to supplement or guide the RGB modality; however, these methods typically adopt asymmetric network structures that rely predominantly on the RGB modality, resulting in limited robustness for all-day detection.
In this paper, we propose EOLO, a novel object detection framework that achieves robust and efficient all-day detection by fusing both RGB and event modalities. Our EOLO framework is built based on a lightweight spiking neural network (SNN) to efficiently leverage the asynchronous property of events. Buttressed by it, we first introduce an Event Temporal Attention (ETA) module to learn the high temporal information from events while preserving crucial edge information.
Secondly, as different modalities exhibit varying levels of importance under diverse lighting conditions, we propose a novel Symmetric RGB-Event Fusion (SREF) module to effectively fuse RGB-Event features without relying on a specific modality, thus ensuring a balanced and adaptive fusion for all-day detection.
In addition, to compensate for the lack of paired RGB-Event datasets for all-day training and evaluation, we propose an event synthesis approach based on the randomized optical flow that allows for directly generating the event frame from a single exposure image.
We further build two new datasets, E-MSCOCO and E-VOC based on the popular benchmarks MSCOCO and PASCAL VOC.
Extensive experiments demonstrate that our EOLO outperforms the state-of-the-art detectors, \eg, RENet~\cite{zhang2023frame}, by a substantial margin ($+3.74\%$ mAP$_{50}$) in all lighting conditions. 
Our code and datasets will be available at \url{https://vlislab22.github.io/EOLO/}.

\end{abstract}

\section{INTRODUCTION}

Object detection is a crucial task that detects the objects of a particular class from digital images and videos, enabling a wide range of applications, \eg, self-driving~\cite{ye2022rope3d,feng2020deep} and robotics~\cite{markovic2014moving, park2020single}. 
In recent years, deep learning-based approaches~\cite{Redmon2018YOLOv3AI,li2022exploring} have demonstrated remarkable effectiveness in the realm of RGB-based object detection.
However, they often suffer from dramatic performance drops when predicting objects under sub-optimal lighting conditions~\cite{rashed2019fusemodnet}, especially under extremely exposed scenes. 

Event cameras, which are bio-inspired sensors that exhibit remarkable advantages over RGB cameras, have recently drawn great attention for self-driving. In particular, event cameras provide rich edge information with high dynamic range (HDR) and high temporal resolution, which can be important in pinpointing the target's location; thus event cameras are beneficial in achieving safe driving, in especially high-speed motion and extremely exposed visual conditions. 
Consequently, some research endeavors, \eg,~\cite{zhu2023visual,zhang2023cmx,zhou2023rgb,shi2023even}, have attempted to fuse RGB and event cameras to augment the performance of downstream tasks. 
These methods typically adopt asymmetric network structures that prioritize RGB features while treating event data as a supplement or guidance. 
For instance, RENet~\cite{zhou2023rgb} proposes a multi-scale RGB-Event fusion framework, where the RGB encoder has significantly larger parameters than that of events. 
However, these methods \textit{fail to balance the importance of event modality} as event cameras can be more advantageous in extreme visual conditions while RGB cameras can be more beneficial in normal conditions. 
The unbalanced fusion consequently results in performance degradation, particularly in extreme exposure conditions (See Tab.~\ref{tab:main_results}).
Therefore, \textit{feature fusion without distinguishing the degree of modality importance may introduce inevitable modality interference, making it challenging to achieve robust all-day object detection}.



\begin{figure}[tbp]
	\centering
	\small
	\includegraphics[width=1\linewidth]{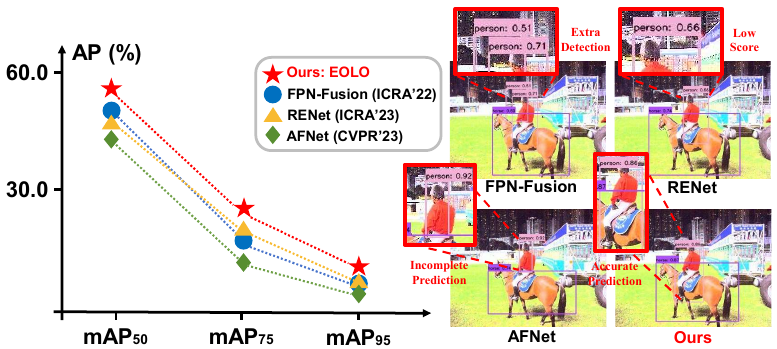}
 \caption{\textbf{Comparison with the SOTA baselines in PASCAL VOC val set under extreme overexposure scenarios.} Our model demonstrates a noticeable improvement compared to other methods in terms of average precision metrics and also exhibits accurate qualitative results.}
	\label{fig:teaser}
\end{figure}

%


\textbf{Motivation:} In this paper, we propose \textbf{EOLO}, a novel event-guided object detection framework that \textit{can achieve robust and efficient all-day object detection by fusing RGB and events while highlighting the modality's importance in different visual conditions}.
Our EOLO integrates a lightweight spiking neural network (SNN) as the feature extractor to better leverage the asynchronous property of event cameras. 
Buttressed by the SNN, we first introduce an event temporal attention (ETA) module to extract the temporal features from events while maintaining crucial edge information (Sec.~\ref{subsec:ETA}). 
Secondly, as different modalities exhibit varying levels of importance under diverse lighting conditions, we propose a novel Symmetric RGB-Event Fusion (SREF) module, to effectively fuse RGB-Event features without relying on a specific modality, thus ensuring balanced and adaptive fusion for all-day detection. 
Our SREF module consists of two key components: 
cross-modality alignment for merging the content and style of both modalities and symmetric modality fusion for balancing the modality fusion (Sec.~\ref{subsec:SREF}).

Additionally, due to the lack of paired RGB-Event datasets for training and evaluation, we thus propose a randomized optical flow-based event synthesis algorithm that derives the corresponding event frame from a single exposure image (Sec.~\ref{subsec:synths}). 
Moreover, based on the well-known benchmarks MSCOCO and PASCAL VOC, we build two synthetic event datasets, E-MSCOCO and E-VOC, to address the scarcity of paired datasets for RGB images and events for all-day conditions. Extensive experiments validate the effectiveness of the proposed approach (see Fig.~\ref{fig:teaser}).

\textbf{Contributions:} In summary, the contributions of our paper are three-fold:
(\textbf{I}) We propose EOLO that combines RGB and event for efficient and robust all-day detection. EOLO consists of an ETA module to learn the temporal features of events and an SREF module to effectively fuse RGB-Event features without relying on a specific modality for a balanced and adaptive fusion for all-day detection. (\textbf{II}) We introduce a randomized optical flow-based event synthesis algorithm that can directly synthesize event frames from a single exposure image and further build two synthetic event datasets, E-MSCOCO and E-VOC. (\textbf{III}) We demonstrate the superiority of EOLO when compared with prior arts with up to \textbf{4\%} increase in average precision (AP) under various exposure conditions. EOLO further exhibits robustness in randomly exposed scenes and real-world scenarios.

\begin{figure*}[t]
\centering
\includegraphics[width=\linewidth]{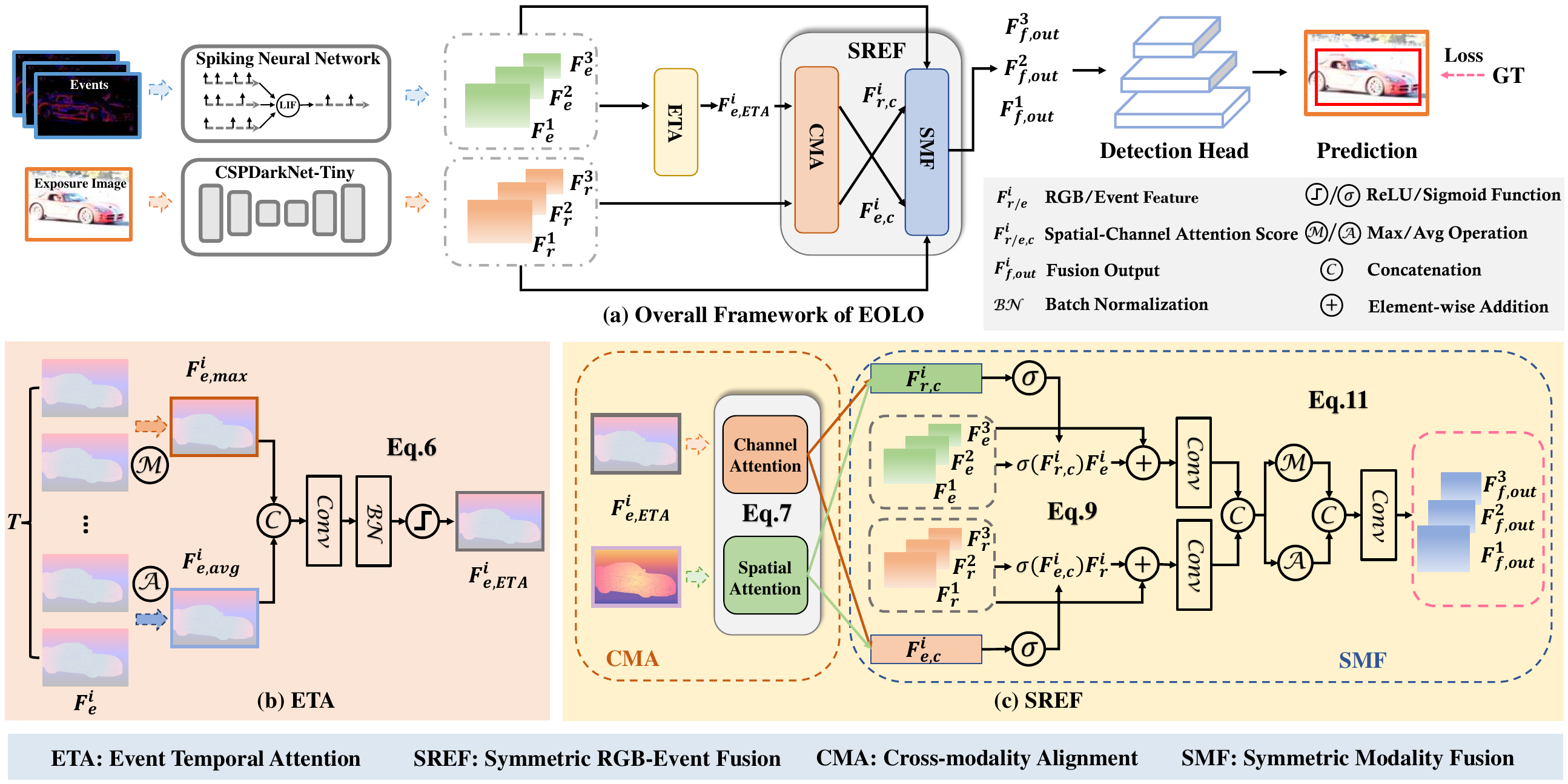}
\caption{(a) The Overall Framework of our proposed EOLO; 
(b) Event Temporal Attention Module (ETA, Sec.~\ref{subsec:ETA}); and (c) Symmetric RGB-Event Fusion Module (SREF, Sec.~\ref{subsec:SREF}), which includes Cross-modality Alignment (CMA) and Symmetric Modality Fusion (SMF). The RGB inputs and event inputs are first processed by the CSPDarknet-Tiny and the Spiking Neural Network to obtain the features of the corresponding modalities $F_r^i$ and $F_e^i$ , respectively. Subsequently, the ETA module extracts and refines the temporal attributes of events, yielding $F_{ETA}^i$. The SREF module then integrates RGB-Event features without relying on a specific modality for a balanced and adaptive fusion. Finally, the fusion features $F_{f,out}^i$ are passed through the detection head to obtain the prediction results. The detection head and the loss function are adapted from~\cite{redmon2018yolov3}. }
\label{fig:framework}
\end{figure*}

\section{RELATED WORK AND PRELIMINARIES}

\noindent\textbf{Event-based Cameras.}
They are bio-inspired sensors, which capture the relative intensity changes asynchronously. In contrast to standard cameras that output 2D images, event cameras output sparse event streams.
When brightness change exceeds a threshold $C$, an event $e_k$ is generated containing position $\textbf{u} = (x,y)$, time $t_k$, and polarity $p_k$:
\begin{equation}
    \Delta L(\textbf{u},t_k) = L(\textbf{u},t_k) - L(\textbf{u},t_k - \Delta t_k) = p_k C.
    \label{eq:event_generate}
\end{equation}

The polarity of an event reflects the direction of the changes (\emph{i.e.}, brightness increase (“ON”) or decrease (“OFF”)). In general, the output of an event camera is a sequence of events, which can be described as: $\mathcal{E} = \{e_k\}_{k=1}^N = \{[\textbf{u}_k, t_k, p_k]\}_{k=1}^N$.
With the advantages of high temporal resolution, high dynamic range, and low energy consumption, event cameras are gradually attracting attention in the fields of tracking~\cite{zhang2021object,zhang2023frame}, identification~\cite{baldwin2022time} and estimation~\cite{nam2022stereo}. For more details, we refer readers to the recent survey papers, \eg,~\cite{gallego2020event, zheng2023deep}.

\noindent\textbf{RGB-Event Fusion for Object Detection.}
Research endeavors have been made in fusing RGB and event cameras for robust objection detection in extreme lighting conditions. The straightforward approaches, \eg,~\cite{zheng2023deep}, transform events into frame-like images, however, this operation discards crucial temporal information of event data. 
Alternatively, RGB and event modality data can be fused together at the feature space. These RGB-Event fusion methods can be categorized into two types based on the fusion stage: middle fusion ~\cite{tomy2022fusing, sun2022event, zhou2023rgb} and late fusion ~\cite{tulyakov2022time, messikommer2022multi}. However, these methods employ asymmetric network structures that prioritize RGB features and treat event data as a supplement, resulting in an imbalance in the importance of the two modalities. Moreover, feature fusion without distinguishing the degree of modality importance and aligning multi-modal features makes it difficult to achieve robust all-day object detection. To address these issues, this study proposes EOLO, which effectively fuses RGB-Event features without relying on a specific modality, thereby ensuring a balanced and adaptive fusion for all-day detection.

\noindent\textbf{Spiking Neural Network (SNN).}
SNNs are potential competitors to artificial neural networks (ANNs) due to their distinguished properties: high biological plausibility, event-driven nature, and low power consumption. 
In SNNs, all information is represented by binary time series data rather than continuous representation, 
leading to significant energy efficiency gains. 
Also,
SNNs possess powerful abilities to extract spatial-temporal features for various tasks, including recognition~\cite{wang2023masked,zhou2022spikformer,deng2022temporal}, tracking~\cite{zhang2022spiking}, segmentation~\cite{kirkland2020spikeseg} and image 
generation~\cite{cao2024spiking}.
In this paper, we adopt the widely used SNN model based on the Leaky Integrate-and-Fire (LIF~\cite{hunsberger2015spiking,burkitt2006review}) neuron, which effectively characterizes the dynamic process of spike generation and can be defined as:
\begin{align}
    & U[n] = e^{\frac{1}{\tau}}V[n-1] + I[n] \label{eq:dis_lif1},\\
    & S[n] = \Theta (U[n] - \vartheta_{\textrm{th}})\label{eq:dis_lif2},\\
    & V[n] = U[n](1-S[n]) + V_{\textrm{reset}}S[n] \label{eq:dis_lif3},
\end{align}
where $n$ is the time step; $U[n]$ is the membrane potential before reset; $S[n]$ denotes the output spike which equals 1 when there is a spike and 0 otherwise; $\Theta(x)$ is the Heaviside step function; $V[n]$ represents the membrane potential after triggering a spike. When the membrane potential exceeds the threshold $\vartheta_{\textrm{th}}$, the neuron will trigger a spike and resets its membrane potential to a value $V_{\textrm{reset}}$ ($V_{\textrm{reset}}<\vartheta_{\textrm{th}}$). The LIF neuron achieves a balance between computing cost and biological plausibility.

\section{The proposed approach: EOLO}

\subsection{Network Overview}
As depicted in Fig.~\ref{fig:framework}, our EOLO framework consists of four major components: the feature extractor (\ie, RGB and event backbones), the Event Temporal Attention Module (ETA), the Symmetric RGB-Event Fusion Module (SREF) and 
the detection predictor for object classification and bounding box regression. EOLO has two distinct branches: the RGB branch and the event branch, which utilize a tiny DarkNet~\cite{redmon2017yolo9000} and a lightweight Spiking ResNet~\cite{zheng2021going}, respectively. Same as YOLOv3~\cite{redmon2018yolov3}, each branch of EOLO produces three different levels of features. 

We choose SNN as the event encoder for its inherent capability to capture spatio-temporal characteristics of events.
For each RGB input $R_i$, we can get the paired event frame $E_i$ through the randomized optical flow-based event synthesis (See Sec.~\ref{subsec:synths}). Then we adopt a commonly used constant coding~\cite{ding2021optimal} (replicating the input for $T$ times) method for $E_i$ as it is proven to have realistic and stable signal simulation capability. 
The encoded event sequence can be processed by: $\{E_0, E_1,...E_T\} = ConstantCode(E_i)$,
where $T$ refers to the time dimension of the event and also the time step of the SNN. Subsequently, the event sequence is fed into the SNN backbone, which generates three distinct levels of event features (\ie, $[F_e^1, F_e^2, F_e^3] = SNN(\{E_0, E_1,...E_T\})$). Similarly, the RGB branch is provided with the corresponding exposure image $R_i$ as input, yielding three multi-level RGB features (\ie, $[F_r^1, F_r^2, F_r^3]$). 

Then, the event features are processed via the ETA module that refines the event features in the temporal dimension while efficiently extracting crucial edge information (Sec.~\ref{subsec:ETA}). 
The event features are then fused with the RGB features via the SREF module that addresses the imbalance of modality importance between event and RGB features, which may introduce inevitable modality interference to achieve all-day object detection (Sec.~\ref{subsec:SREF}).

After obtaining the fusion features through the fusion module, we ultimately feed these features into the Feature Pyramid Network (FPN) and detection head for classifying and detecting targets by following the design of YOLOv3. We now describe the technical details.

\subsection{Event Temporal Attention (ETA)}
\label{subsec:ETA}
The ETA module is proposed to extract the temporal attributes from events while preserving crucial edge information, enabling improved integration with RGB features.
Since the event features $F_e^i \in \mathbb{R}^{C \times T\times H \times W}$ obtained from the SNN are stored as spikes, \ie, 0/1 at each position. We first need to perform a de-discretization process. Meanwhile, the information of $F_e^i$ at time dimension needs to be filtered. Inspired by~\cite{woo2018cbam}, we extract the max and average features within the time dimension: $\{F_{e,max}^i, F_{e,avg}^i\} =  \{\mathop{max}^{T}(F_{e}^i), \mathop{avg}^{T}(F_{e}^i)\}$. 
Then, we design a temporal attention (TA) to aggregate features from temporal perspective:
\begin{align}
    F_{e,ETA}^i &= TA([F_{e,max}^i, F_{e,avg}^i)]),\\
     &= \sigma(BN(\psi{_1}([F_{e,max}^i, F_{e,avg}^i)]))),
\end{align}
where $i$ denotes the feature level; $[\cdot]$ denotes channel-wise concatenation; $\psi{_k}$ denotes the convolution operation where kernel size is $k\times k$; $\sigma$ and $BN$ denote the sigmoid function and batch normalization.

\subsection{Symmetric RGB-Event Fusion (SREF)}
\label{subsec:SREF}
The SREF module is designed to effectively integrate RGB-Event features without depending on a specific modality, thus ensuring a balanced and adaptive fusion that facilitates all-day detection purposes.
SREF contains two components: (1) Cross-modality Alignment (CMA), and (2) Symmetric Modality Fusion (SMF). CMA emphasizes the motion cues from both channel and spatial dimensions, and SMF maintains a balanced fusion between the event and RGB features in a symmetric manner, thus preventing the loss of crucial modal information.  


\noindent
\textbf{CMA.} Given event modality features $F_{e,ETA}^i \in \mathbb{R}^{C \times H \times W}$, we calculate the channel and spatial attention scores to obtain the fusion features by: 
\begin{align}
    F_{e,s}^i &= \mathcal{R}^{(1, C,HW)}(F_{e,ETA}^i)\times  \mathcal{R}^{(1, HW,1)}(\mathcal{S}(\psi{_1}(F_{e,ETA}^i))),\nonumber\\
    F_{e,c}^i &=  \sigma(\psi{_1}(\psi{_1}(\mathcal{R}^{(C,1,1)}(F_{e,s}^i))))F_{e,ETA}^i ,
\end{align}
where $F_{e,s}^i$ and $F_{e,c}^i$ are event features enhanced in the spatial and channel dimensions at level $i$, respectively. $\mathcal{S}$ denotes the softmax function and $\mathcal{R}(\cdot)$ is a reshape function with a target shape~$(\cdot)$. $F_{e,s}^i$ extracts detailed and fine-grained information from the spatial dimension, then dives into the channel dimension to obtain finer channel-wise feature $F_{e,c}^i$.
However, existing RGB-Event fusion approaches~\cite{zhu2023visual,zhang2023cmx,zhou2023rgb} usually favor RGB feature extraction, while fusing event features with RGB features as auxiliary elements. A common fusion method is formulated as follows:
\begin{align}
    F_{fusion}^i &=  \sigma(F_{e,c}^i)F_r^i  + F_r^i.
    \label{eq:basicfusion}
\end{align}
We denote such a fusion module as $basic~fusion$. \textit{This destroys the balance between the two modalities, and the RGB identity at the end of Eq.~\ref{eq:basicfusion} makes the RGB features largely preserved, which in turn leads to the whole fusion module being RGB-dominated and event-assisted}. Our experiments (See Tab.~\ref{tab:component}) demonstrate the performance degradation with basic fusion under extremely exposed scenes.

\noindent\textbf{SCF.} To balance the modality importance between the RGB-Event modalities, the SCF simultaneously processes the feature embeddings of the two modalities and combines them in a symmetric manner:
\begin{align}
    F_{fusion}^i &=  (\sigma(F_{e,c}^i)F_r^i  + F_r^i) + (\sigma(F_{r,c}^i)F_e^i  + F_e^i),\\
    F_{f,max}^i &= max^C[\psi{_1}(F_{fusion,r}^i), \psi{_1}(F_{fusion,e}^i)],
    \label{eq:SCF}
\end{align}
where $max^C$ denotes the max operation in dimension $C$; 
$F_{fusion,{r}}^i$ means the RGB part ($\sigma(F_{e,c}^i)F_r^i  + F_r^i$) and $F_{fusion,{e}}^i$ means the event part ($\sigma(F_{r,c}^i)F_e^i  + F_e^i$) of the fusion feature.

Finally, we refine the fusion features by
concatenating global average features and max features, which represent the salient features in order to get cross-correlation output containing strong cues and refined information:
\begin{align}
    F_{f,out}^i &= \psi{_3}([F_{f,max}^i, F_{f,avg}^i]).
    \label{eq:SMF_maxavg}
\end{align}
Further ablation study (See Tab.~\ref{tab:component}) demonstrates that the fusion output by jointly combining $max$ and $avg$ features yields superior performance in downstream tasks compared to the fusion output derived from using either $max$ or $avg$ features individually.
Inheriting from YOLOv3~\cite{redmon2018yolov3}, $[F_{f,out}^1, F_{f,out}^2, F_{f,out}^3]$ are fed into the classifier and regressor to locate the target. 
We adopt the YOLOv3's loss function which contains three parts: 
object loss using MSEwithLogit, class loss using CrossEntropy, and bounding box loss using the intersection of union (IOU) metric.

\begin{figure}[tbp]
	\centering
	\small
  \includegraphics[width=1\linewidth]{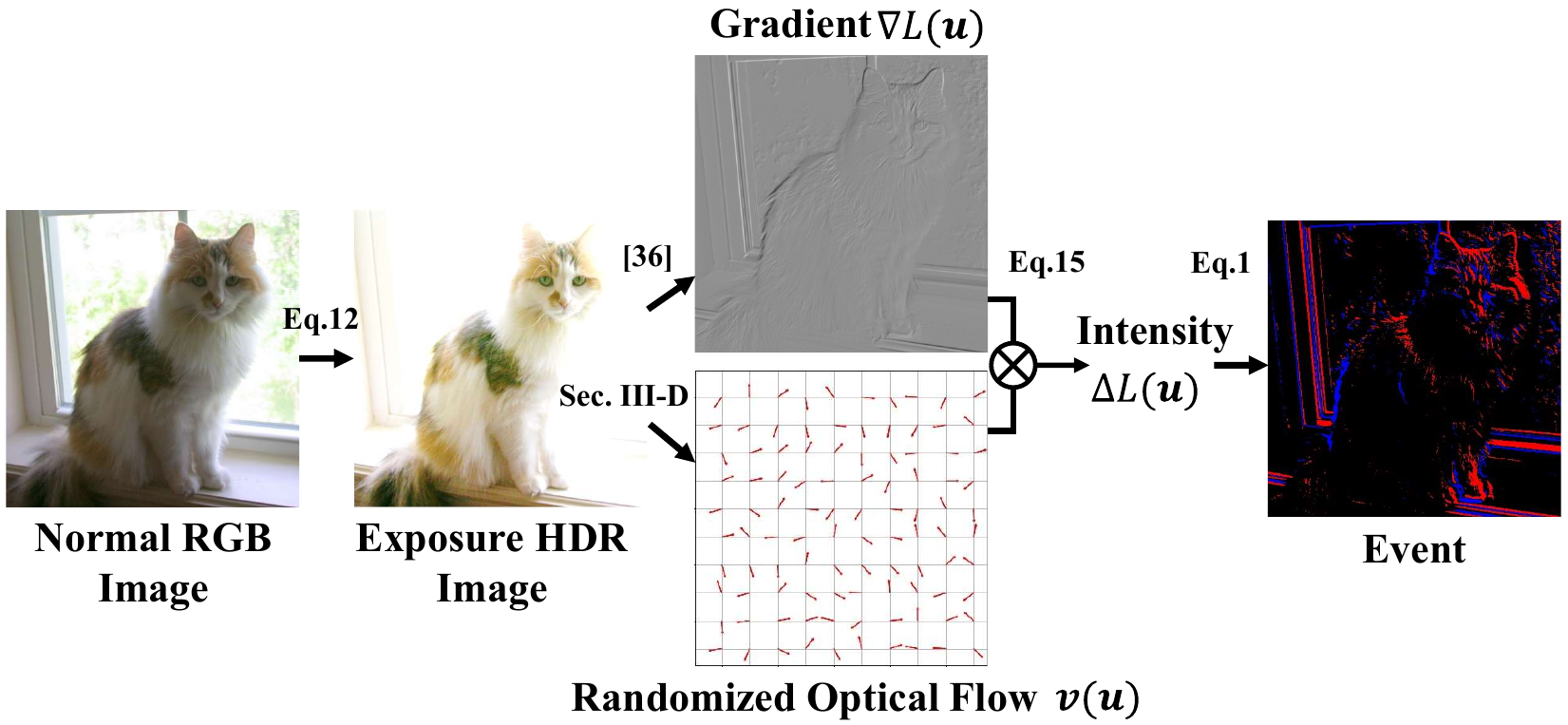}
	\caption{\textbf{Overview of our randomized optical flow-based event synthesis algorithm.} `$\otimes$' denotes the dot product. } 
	\label{fig:synthesis}
\end{figure}

\subsection{Randomized Optical Flow-based Event Synthesis}
\noindent\textbf{Extreme-exposure Data Transformation.}
\label{subsec:extreme_trans}
As there is a lack of all-day detection datasets, we leverage the normal-light RGB dataset and synthesize the extremely exposed dataset by brightness transformation. 
Specifically, we first transform the image from the RGB to the HSV color space. Then, an exposure factor $\alpha$ is multiplied with the `$V$' component of the HSV representation for exposure processing:
\begin{equation}
    V_{exp} = V * \alpha.
    \label{eq:exposure}
\end{equation}
When $\alpha<0$, the resulting image emulates underexposure, and conversely, overexposure is simulated for $\alpha$ greater than zero. This way, it can directly modulate the brightness values, which are predominantly accountable for the perception of over- or under-exposed scenes.

\noindent\textbf{Event Synthesis based on Randomized Optical Flow.}
\label{subsec:synths}
To obtain paired event data, we propose a novel event frame synthesis method that generates event frames by the randomized optical flow and luminance gradients. Only a single RGB/HDR image is required to generate the corresponding event frames. Notably, we assume that optical flow exists at every position, allowing this approach to accurately \textit{simulate events captured by a moving event camera}.


In a small time interval, the brightness consistency assumption~\cite{horn1981determining} is conformed, under which the intensity change in a vicinity region remains the same. By using Taylor's expansion, we can approximate intensity change by:
\begin{align}
    \Delta L(\textbf{u}, t) &= L(\textbf{u}, t) - L(\textbf{u}, t-\Delta t), \\
    &= \frac{\delta L}{\delta t}(\textbf{u}, t)\Delta t + O(\Delta t^2)\approx \frac{\delta L}{\delta t}(\textbf{u}, t)\Delta t,
\end{align}
where $\textbf{u}=(x,y)$ denotes the position.
Substituting the brightness constancy assumption ($\frac{\delta L}{\delta t}(\textbf{u}(t),t) + \nabla L(\textbf{u}(t),t) \cdot \textbf{v}(\textbf{u})) = 0.$) into the above equation, we can obtain:
\begin{equation}
    \Delta L(\textbf{u}) \approx -\nabla L(\textbf{u}) \cdot \textbf{v}(\textbf{u})\Delta t, \label{eq:bright}
\end{equation}
which indicates that the brightness changes are caused by intensity gradients $\Delta L = (\frac{\delta L}{\delta x},\frac{\delta L}{\delta y})$
moving with velocity $\textbf{v}(\textbf{u})$ over a displacement $\Delta \textbf{u} =\textbf{v}\Delta t $. As expressed by the dot product in Eq.~\ref{eq:bright}, if
the moving direction is parallel to the brightness gradient (\ie, $\textbf{v}\perp \nabla L$), no events are generated. 

Next, we consider the generation of random velocity vectors $\textbf{v}(\textbf{u})$. For each vector, $\textbf{v}_i = (v_i^x, v_i^y)$, we assign them with randomized optical flow:
$(v_i^x, v_i^y) = (cos(\theta), sin(\theta))$
, where $\theta $ is randomly sampled from $[-\pi, \pi]$, controlling the direction of optical flow.
Based on the velocity matrix $\textbf{v}(\textbf{u})$, we could obtain the correspondent gradient matrix $\nabla L(\textbf{u}) $ by the Sobel method~\cite{sobel19683x3}. With $\textbf{v}(\textbf{u})$ and $\nabla L(\textbf{u}) $, we can get $\Delta L(\textbf{u}) $ to generate event data with Eq.~\ref{eq:event_generate}. The pipeline of our randomized optical flow-based event synthesis algorithm is depicted in Fig.~\ref{fig:synthesis}.

\begin{figure*}[htbp]
	\centering
	\small
 \begin{tabular}{@{}c@{}}
\includegraphics[width=1\linewidth]{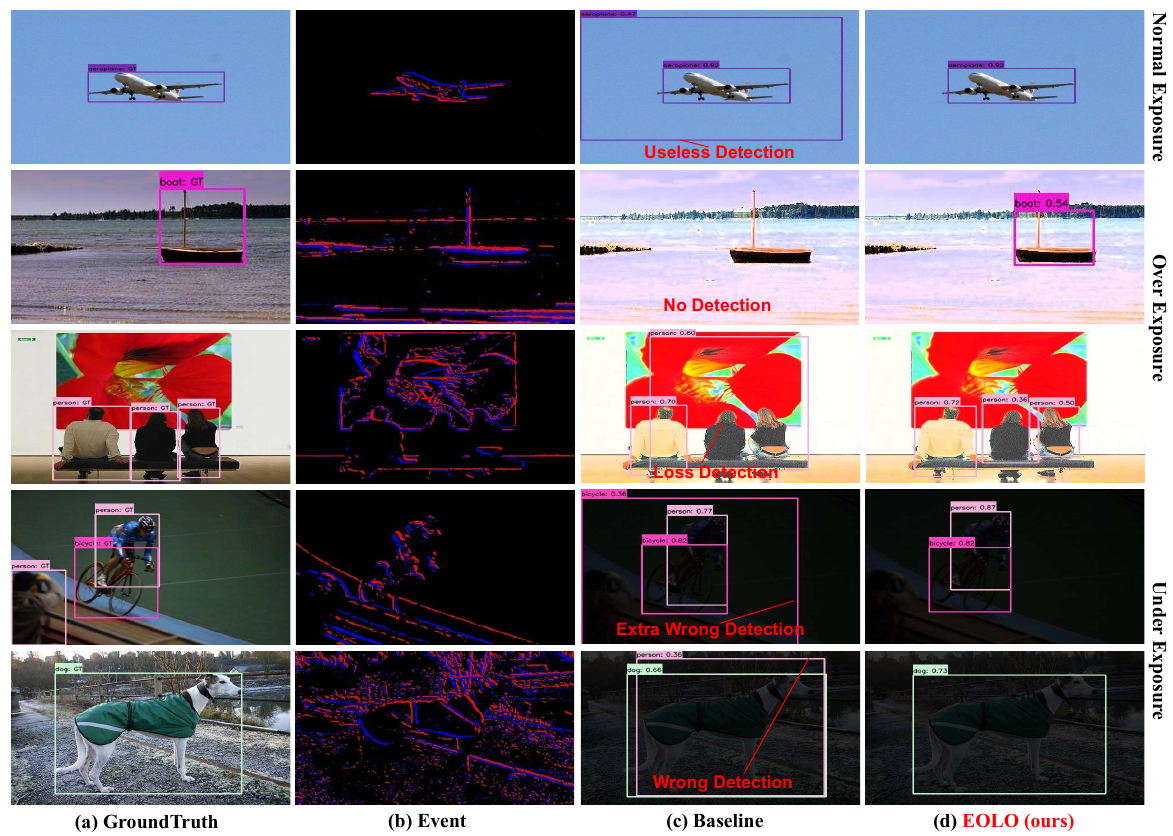}\\
\end{tabular}
	\caption{\textbf{Qualitative comparison of our EOLO on the PASCAL VOC  dataset under all-day exposure conditions.}} 
 \label{fig:main_vis}
	\label{fig:results}
	\vspace{-0.4cm}
\end{figure*}

\section{Experiments}

\subsection{Datasets and Baselines}
To simulate extreme exposure scenarios, we first obtain the exposure dataset (EXPOSE-VOC and EXPOSE-COCO) through the exposure transformation algorithm (Sec.~\ref{subsec:extreme_trans}) based on two well-known detection datasets: PASCAL VOC~\cite{everingham2015pascal} and MSCOCO 2017~\cite{lin2014microsoft}. For each exposure dataset, we set up extreme underexposure ($\alpha$=0.2) and extreme overexposure ($\alpha$=5.0) conditions, respectively. As mentioned in Sec.~\ref{subsec:synths}, we then use the randomized optical flow event synthesis method to generate the paired event frames datasets: E-VOC and E-COCO. 
We use YOLOv3-Tiny as the single modality baseline and compare our results with the state-of-the-art RGB-Event fusion methods: FPN-Fusion~\cite{tomy2022fusing} (ICRA'22), RENet~\cite{zhou2023rgb} (ICRA'23) and AFNet~\cite{zhang2023frame} (CVPR'23). We re-implement these methods by replacing the fusion module and setting the event backbone as ResNet-18 for fair comparisons.

\subsection{Experimental Settings}
\noindent\textbf{Implementation Details.}
We choose Spiking ResNet-18~\cite{zheng2021going} as the event backbone and CSPDarknet-Tiny~\cite{bochkovskiy2020yolov4} as the RGB backbone. We use YOLO's common data augmentation methods for RGB inputs, including color jitter, random crop, and random flip. Events are not augmented. Both RGB and events are resized to 320$\times$320. The time step of SNN is set to 4. We adopt the SGD optimizer and set the initial learning rate as 5e-4, along with the step learning rate scheduler. All models are trained for 50 epochs with batch size 32. 

\noindent\textbf{Evaluation Metrics.}
We use the most commonly used mean average precision (mAP) as our evaluation metric,  where AP$_{50}$ and AP$_{75}$ denote the intersection of union (IOU) threshold is set to 0.5 and 0.75, respectively.

\begin{table}[t]
\renewcommand\arraystretch{1.2}
\caption{
\textbf{Comparison with SOTA RGB-Event fusion methods.}
}
\vspace{-6mm}
\label{tab:main_results}
\begin{center}
\setlength{\tabcolsep}{0.5mm}
\resizebox{\linewidth}{!}{
\begin{tabular}{l|l|c|cccc}
\hline

\hline
 \multirow{2}{*}{\textbf{Exposure}}& \multirow{2}{*}{\textbf{Method}} &\multirow{2}{*}{\textbf{Modality}}&\multicolumn{2}{c}{\textbf{EXPOSE-VOC}}&\multicolumn{2}{c}{\textbf{EXPOSE-COCO}}\\
 && &AP$_{50}$ (\%) & AP$_{75}$ (\%) & AP$_{50}$ (\%) & AP$_{75}$ (\%) \\
\hline
\multirow{6}{*}{Normal} & 
RGB only & \multirow{2}{*}{Single} & 58.29 & 29.62 & 33.90 &16.40\\
& Event only &  & 12.33  & 3.35 & 9.25 & 2.49 \\ \cline{2-7}
&FPN-Fusion\cite{tomy2022fusing}& \multirow{4}{*}{RGB-E} & 58.55 &  30.93 & 32.15 & 16.32 \\
& AFNet\cite{zhang2023frame} && 58.03 &31.09 & 34.33& 17.74 \\
&RENet\cite{zhou2023rgb} & & 59.33 & 31.69 & 35.57 & 17.84\\
& \textbf{EOLO (ours)} & & \textbf{61.29} & \textbf{33.26} & \textbf{37.58}&\textbf{18.20}  \\ \hline
\multirow{6}{*}{Under} & 
RGB only & \multirow{2}{*}{Single} & 56.08 & 28.81 & 30.20 & 14.37\\
& Event only &  & 13.25 & 3.12 & 9.32 & 2.56 \\ \cline{2-7}
&FPN-Fusion\cite{tomy2022fusing} & \multirow{4}{*}{RGB-E} & 56.83 & 29.35 & 31.01 & 14.98\\
& AFNet\cite{zhang2023frame}  &  & 55.60 & 28.93 &32.45 & 15.15  \\
&RENet\cite{zhou2023rgb} &  & 57.42 & 31.33 &33.62 & 16.01 \\
& \textbf{EOLO (ours)} &  & \textbf{60.61} & \textbf{32.70} & \textbf{35.24}&\textbf{17.59} \\
\hline
\multirow{6}{*}{Over} & 
RGB only & \multirow{2}{*}{Single} & 52.66  & 22.13 & 28.94 & 13.09\\
& Event only &  & 11.47 & 2.89 & 8.71 & 2.14 \\ \cline{2-7}
&FPN-Fusion\cite{tomy2022fusing}& \multirow{4}{*}{RGB-E} & 52.21 & 26.69 & 29.28 & 13.45\\
& AFNet\cite{zhang2023frame} &  & 51.52 & 26.14 &  30.92 & 13.98  \\
&RENet\cite{zhou2023rgb} &  & 52.13 & 27.70 &  31.89 & 14.67\\
& \textbf{EOLO (ours)} &  & \textbf{55.87} & \textbf{28.97} & \textbf{33.10} & \textbf{15.21}  \\
\hline

\hline
\end{tabular}
}
\end{center}
\vspace{-3mm}
\end{table}

\subsection{Main Results}
To demonstrate the effectiveness of our method under all-day detection scenes, we evaluate EOLO under three exposure conditions: Normal Exposure, Under Exposure, and Over Exposure. Tab.~\ref{tab:main_results} shows the evaluation results on the PASCAL VOC and MSCOCO datasets compared with the state-of-the-art fusion methods. In particular, our proposed EOLO achieves 61.29\% and 37.58\% AP$_{50}$ in normal exposed VOC and COCO, respectively, yielding a substantial improvement compared with the single modality-based YOLO baseline. For challenging extreme exposure scenarios of VOC dataset, EOLO obtains 60.61\% and 55.87\% AP$_{50}$ in underexposure and overexposure conditions, outperforming the runner-up by 3.19\% and 3.74\%, respectively. Fig.~\ref{fig:main_vis} presents a qualitative comparison between our approach and the RGB baseline. Notably, the performance of the RGB detector experiences a significant decline in extreme exposure scenes. In contrast, EOLO is capable of achieving more robust and accurate detection in all-day exposure scenarios.

\subsection{Ablation Study}
\noindent\textbf{Exposure Factor in Randomized Flow-based Event Synthesis.}
First, we explore the effect of the exposure factor $\alpha$ (in Eq.~\ref{eq:exposure}) in our experiments. As illustrated in Table~\ref{tab:expose_factor}, the performance of the higher exposure levels tends to deteriorate. This can be attributed to the gradual decline in effective information as the exposure level escalates. Secondly, we conduct our model under the random exposure condition. In comparison to the challenging full over- or underexposure scenario, EOLO demonstrates enhanced performance in the mixed exposure case, substantiating the model's robustness in intricate environments. Additionally, we observe that employing the randomized optical flow (OF) results in superior performance compared to the fixed optical flow manner.

\noindent\textbf{Impact of SNN Time Steps and Computational Cost.}
To further emphasize the low-energy nature of our SNN backbone, we perform a comparative analysis of the AP and energy consumption between the SNN and its corresponding ANN model. The energy consumption calculation is from~\cite{yao2023attention}. As shown in Tab.~\ref{tab:energy}, when the time step is 1, EOLO presents significantly lower energy consumption, amounting to merely 96\% of that exhibited by the ANN model. In addition, the model performance improves when the number of time steps increases, indicating that our model can effectively minimize energy consumption while maintaining competitive performance.

\noindent\textbf{Impact of Modules.}
Our proposed module has two key components: ETA and SREF, where SREF includes Cross-modality Alignment (CMA), and Symmetric Modality Fusion (SMF). Tab.~\ref{tab:component} reveals that all of our proposed modules contribute to the detection performance of EOLO.
(1) SREF: We can opt to utilize either SMF$_{avg}$ or SMF$_{max}$, or a combination of both (Eq.~\ref{eq:SMF_maxavg}). The results by only using $basic fusion$ are suboptimal and even inferior to those achieved with RGB modalities alone. Furthermore, using SMF$_{max}$ module leads to an improvement of 3.51\% and 1.91\% in AP$_{50}$ and AP$_{75}$, respectively. This demonstrates the significance of maintaining a balance between the two modalities during cross-alignment. 
(2) ETA: The ETA module contributes to the model attaining state-of-the-art performance, achieving 60.61\% and 32.70\% in AP$_{50}$ and AP$_{75}$, respectively, thereby exemplifying its effectiveness.

\begin{table}[t]
\renewcommand\arraystretch{1.1}
\begin{center}
\caption{\textbf{Ablation study on exposure factor of randomized flow-based event synthesis algorithm.} }
\label{tab:expose_factor}
\begin{tabular}{c|c|c|cc}
\hline
\hline
 Condition  &Factor $\alpha$ & Random OF& AP$_{50}$ (\%)  &  AP$_{75}$ (\%)  \\
\hline
\multirow{4}{*}{\shortstack{Under Exposure\\ ($\alpha \downarrow$, expose $\uparrow$)}} & \multirow{2}{*}{1/4} & & 60.52 & 32.39  \\
& & \checkmark &\textbf{60.67} & \textbf{32.80}\\ 
 & \multirow{2}{*}{1/5} & & 60.17 & 32.24  \\
& & \checkmark & 60.61 & 32.70 \\ \hline
\multirow{4}{*}{\shortstack{Over Exposure\\ ($\alpha \uparrow$, expose $\uparrow$)}} & \multirow{2}{*}{4} &  & 56.83 & 29.72 \\
& & \checkmark &\textbf{57.06} &\textbf{29.83} \\ 
 & \multirow{2}{*}{5} &  & 55.70& 27.63 \\
& & \checkmark & 55.87 &28.97 \\\hline
\multirow{2}{*}{Random Exposure} & \multirow{2}{*}{[$\frac{1}{5},5$]} &  & \textbf{57.21} & 29.29 \\
& & \checkmark & 57.19& \textbf{29.84}\\ 
\hline
\hline
\end{tabular}
\end{center}
\end{table}
\begin{table}[t]
\renewcommand\arraystretch{1.1}
\begin{center}
\caption{\textbf{Ablation study on SNN Time Steps and Energy.} }
\label{tab:energy}
\begin{tabular}{ccccc}
\hline

\hline

Backbone& Time Step & Energy (mJ)&  AP$_{50} (\%)$  & AP$_{75}$ (\%) \\
\hline
ResNet-18&/ & 16.73  & 56.83 & 29.35\\ \hline
\multirow{4}{*}{\shortstack{Spiking \\ResNet-18}}&1 &0.78 (-96\%) & 59.66 & 31.70\\
&2 & 0.86 (-95\%) & 59.90 & 31.64 \\
&4 & 1.27 (-93\%)&60.61 & 32.70 \\
&8 & 2.19 (-87\%) &\textbf{61.11} &\textbf{33.71}\\
\hline

\hline
\end{tabular}
\end{center}
\end{table} 

\begin{table}[t]
\renewcommand\arraystretch{1.1}
\begin{center}
\caption{\textbf{Ablation study on key components on EXPOSE-VOC in underexposure condition.} }
\label{tab:component}
\begin{tabular}{ccccc|cc}
\hline

\hline
 basic  & CMA & SCF$_{max}$ &SCF$_{avg}$ & ETA &  AP$_{50}$ (\%)  &  AP$_{75}$ (\%)  \\
\hline
\checkmark & & & & & 55.22& 26.54 \\
&\checkmark & & & & 56.12& 29.31 \\
& \checkmark&\checkmark & & &59.63 & 31.22 \\
& \checkmark& \checkmark  & \checkmark& &60.16 & 32.45 \\
& \checkmark& \checkmark & \checkmark &\checkmark  & \textbf{60.61} & \textbf{32.70} \\

\hline

\hline
\end{tabular}
\end{center}
\end{table}

\begin{figure}[t]
	\centering
  \includegraphics[width=1\linewidth]{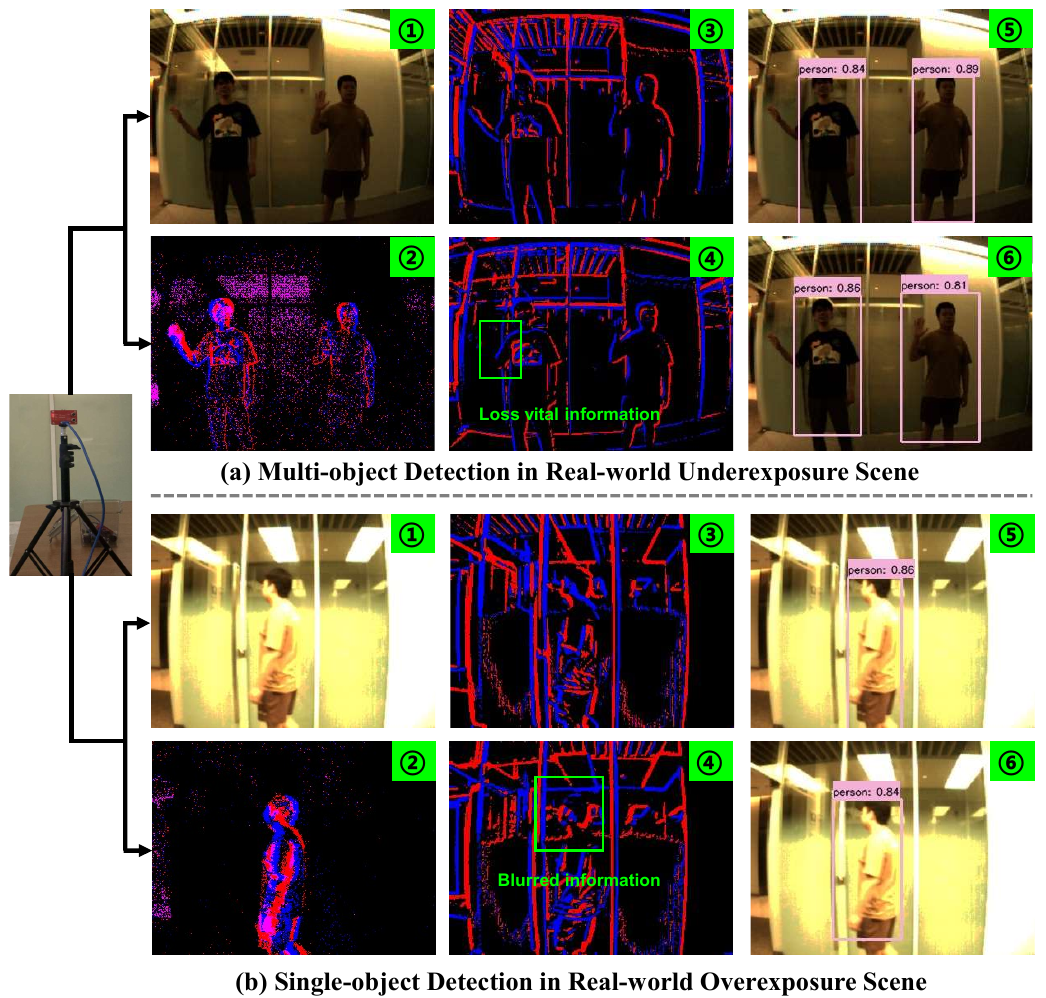}
	\caption{\textbf{Visualization of real-world detection under extreme (a) underexposure scenarios, and (b) overexposure scenarios. } \ding{172} and \ding{173} denote real RGB image and real event, respectively, captured by a DAVIS-346 event-based camera. By event synthesis method (Sec.~\ref{subsec:extreme_trans}), we can obtain the randomized optical flow-based event \ding{174} and its fixed counterpart \ding{175}. In real-world scenes, EOLO yields excellent detection results with inputs of paired RGB-real events (\ding{176}) or paired RGB-synthetic events (\ding{177}).}
	\label{fig:real_world_process}
\end{figure}

\subsection{Evaluation on Real-world Scenes}
We also evaluate the effectiveness of our model in real-world scenarios. The video is captured by a DAVIS-346 event-based camera, which equips a 346$\times$260 pixels dynamic vision sensor (DVS) and an active pixel sensor (APS). 
It can simultaneously provide events and aligned RGB images of a scene. 
As depicted in Fig.~\ref{fig:real_world_process}, we employ real exposure images to generate the corresponding synthetic events. Notably, our algorithm exhibits more contour information comparing to the real event, 
since our approach accurately simulates the event captured by an event camera in a motion situation (Sec.~\ref{subsec:synths}), whereas the real event is captured by event cameras at rest.
In addition, the randomized optical flow-based event exhibits enhanced robustness while preventing the information loss that might occur in comparison to its fixed counterpart. 
Finally, EOLO detects the objects based on the paired RGB-synthetic/real event and obtains accurate detection results.

\section{CONCLUSIONS}
In this work, we proposed a multi-modality framework for all-day exposure object detection with RGB-Event inputs. Our approach incorporated a bio-inspired spiking neural network (SNN) for capturing spatial-temporal features of events. 
Then, we introduced an event temporal attention (ETA) module to refine the temporal features from events, and designed a novel symmetric RGB-Event fusion (SREF) module to effectively achieve cross-modal alignment, ensuring a balanced and robust all-day detection.
In addition, to address the lack of paired RGB-Event datasets, we proposed a randomized optical flow-based event synthesis algorithm capable of generating the corresponding event
frame from a single exposure image. We further built and publicly released two new datasets E-MSCOCO and E-VOC. Extensive experiments demonstrated our EOLO achieves state-of-the-art performance under various challenging exposure conditions. EOLO further revealed its robustness in randomly exposed scenes and real-world scenarios. 

In the future, we aim to apply EOLO in more 
adverse lighting conditions and explore its implementation on robots or drones for more real-world detection applications. Meanwhile, we plan to optimize and enhance our event synthesis algorithm to generate higher-quality RGB-Event datasets to promote further research.

\clearpage
{\small
\bibliographystyle{IEEEtran}
\bibliography{IEEEexample}
}

\end{document}